\def\paperlanguage{}
\pdfoutput=1 

\documentclass[letterpaper, 10 pt, conference]{ieeeconf}  

\usepackage{bm}
\usepackage{cite}
\usepackage{flushend}
\include{preamble}

\newcommand{\ctext}[1]{\raise0.2ex\hbox{\textcircled{\scriptsize{#1}}}}

\IEEEoverridecommandlockouts                              

\title{\LARGE \textbf
  {
    \switchlanguage%
    {%
      Robotic State Recognition with Image-to-Text Retrieval Task of Pre-Trained Vision-Language Model and Black-Box Optimization
    }%
    {%
      ブラックボックス最適化と事前学習済み大規模視覚-言語モデルのITRタスク基づく\\ロボットの二値状態認識
    }%
  }
}

\author{Kento Kawaharazuka$^{1}$, Yoshiki Obinata$^{1}$, Naoaki Kanazawa$^{1}$, Kei Okada$^{1}$, and Masayuki Inaba$^{1}$
  \thanks{$^{1}$ The authors are with the Department of Mechano-Informatics, Graduate School of Information Science and Technology, The University of Tokyo, 7-3-1 Hongo, Bunkyo-ku, Tokyo, 113-8656, Japan.
    {\texttt\small [kawaharazuka, obinata, kanazawa, k-okada, inaba]@jsk.t.u-tokyo.ac.jp}
  }
}
\begin{document}

\maketitle
\thispagestyle{empty}
\pagestyle{empty}

\begin{abstract}
  \switchlanguage%
  {%
    State recognition of the environment and objects, such as the open/closed state of doors and the on/off of lights, is indispensable for robots that perform daily life support and security tasks.
    Until now, state recognition methods have been based on training neural networks from manual annotations, preparing special sensors for the recognition, or manually programming to extract features from point clouds or raw images.
    In contrast, we propose a robotic state recognition method using a pre-trained vision-language model, which is capable of Image-to-Text Retrieval (ITR) tasks.
    We prepare several kinds of language prompts in advance, calculate the similarity between these prompts and the current image by ITR, and perform state recognition.
    By applying the optimal weighting to each prompt using black-box optimization, state recognition can be performed with higher accuracy.
    Experiments show that this theory enables a variety of state recognitions by simply preparing multiple prompts without retraining neural networks or manual programming.
    In addition, since only prompts and their weights need to be prepared for each recognizer, there is no need to prepare multiple models, which facilitates resource management.
    It is possible to recognize the open/closed state of transparent doors, the state of whether water is running or not from a faucet, and even the qualitative state of whether a kitchen is clean or not, which have been challenging so far, through language.
  }%
  {%
    日常生活支援や警備タスクを行うロボットの認識行動において, ドアの開閉状態や電気のオンオフなど, 環境や物体の状態認識は欠かせない.
    これまでこれらの状態認識は, 点群や画像から手動のプログラミングにより特徴を取り出す, 手動アノテーションからニューラルネットワークを学習させる, 特別なセンサを用意して認識するなどの方法が取られてきた.
    これに対し本研究では, Image-to-Text Retrieval (ITR)タスクが可能な事前学習済み視覚-言語モデルを用い, ロボットのための状態認識手法を提案する.
    事前に数種類のプロンプトを用意し, 現在画像についてCLIPにより類似度を計算, これを用いて二値状態認識を行う.
    この際, 遺伝的アルゴリズムによりそれぞれのプロンプトに対して最適な重み付けを行うことで, より高い精度で状態認識が可能となる.
    また, 認識器ごとにプロンプトとそれらの重みのみ用意すれば良いため, モデルを複数用意する必要がなくリソースのマネジメントが容易になる.
    本研究によりニューラルネットワークの再学習や手動のプログラミングを必要とせず, 複数のプロンプトを用意するのみで多様な状態認識が可能となること, また, これまで難しかった水の認識や透明なドアの開閉認識, 綺麗さなどの質的な状態の認識まで可能になることを実験から示す.
  }%
\end{abstract}

\section{INTRODUCTION}\label{sec:introduction}
\switchlanguage%
{%
  Robots are expanding their range of activities not only to fixed environments such as factories and laboratories, but also to human living spaces, nursing care facilities, and disaster sites \cite{okada2005daily, saito2011subwaydemo, klamt2018centaur}.
  For robots that provide daily life support, security, and nursing care, the state recognition of the environment and objects is indispensable.
  In particular, binary state recognition, e.g. open/closed state of doors and on/off state of lights is most frequently used, which cause the motion switching of robots.
  So far, this state recognition has been done by human programming to extract features from point clouds or raw images \cite{borgsen2014door, quintana2018door}, by manually annotating datasets and training neural networks \cite{li2020modifiedyolov3}, or by preparing a special sensor for the state to be recognized \cite{takahata2020coaxial}.
  On the other hand, these methods require programming, annotation, training, or new sensors, which are expensive to construct for various state recognitions.
}%
{%
  今日のロボットの行動範囲は, 工場や実験室などの定まった環境だけでなく, 人間の住む生活の場や介護現場, 災害現場などへと広がりつつある\cite{okada2005daily, saito2011subwaydemo}.
  また, 日常生活支援や警備, 介護支援等を行うロボットにおいて, 環境や物体の状態認識は欠かせない.
  特に, ドアの開閉や電気のオンオフなど, ロボットの動作分岐が生じる二値状態認識は最も多用されている.
  この際, 透明なドアの開閉状態や蛇口からの出水状態, バッグの口の開閉状態やキッチンが綺麗かどうかといった質的な状態など, 認識したい状態は多様である.
  これまでこれらの二値状態認識は, 点群や生画像から人間がプログラミングにより特徴を取り出す\cite{borgsen2014door, quintana2018door}, 手動アノテーションを行ったデータセットを集めニューラルネットワークを学習させる\cite{li2020modifiedyolov3}, その状態認識に特別なセンサを用意して認識する\cite{takahata2020coaxial}などの方法が取られてきた.
  一方で, これには認識したい状態に応じたプログラミングやアノテーションと学習, 新しいセンサの取り付けが必要であり, 多様な状態認識を構築するには大きなコストがかかる.
}%

\begin{figure}[t]
  \centering
  \includegraphics[width=0.8\columnwidth]{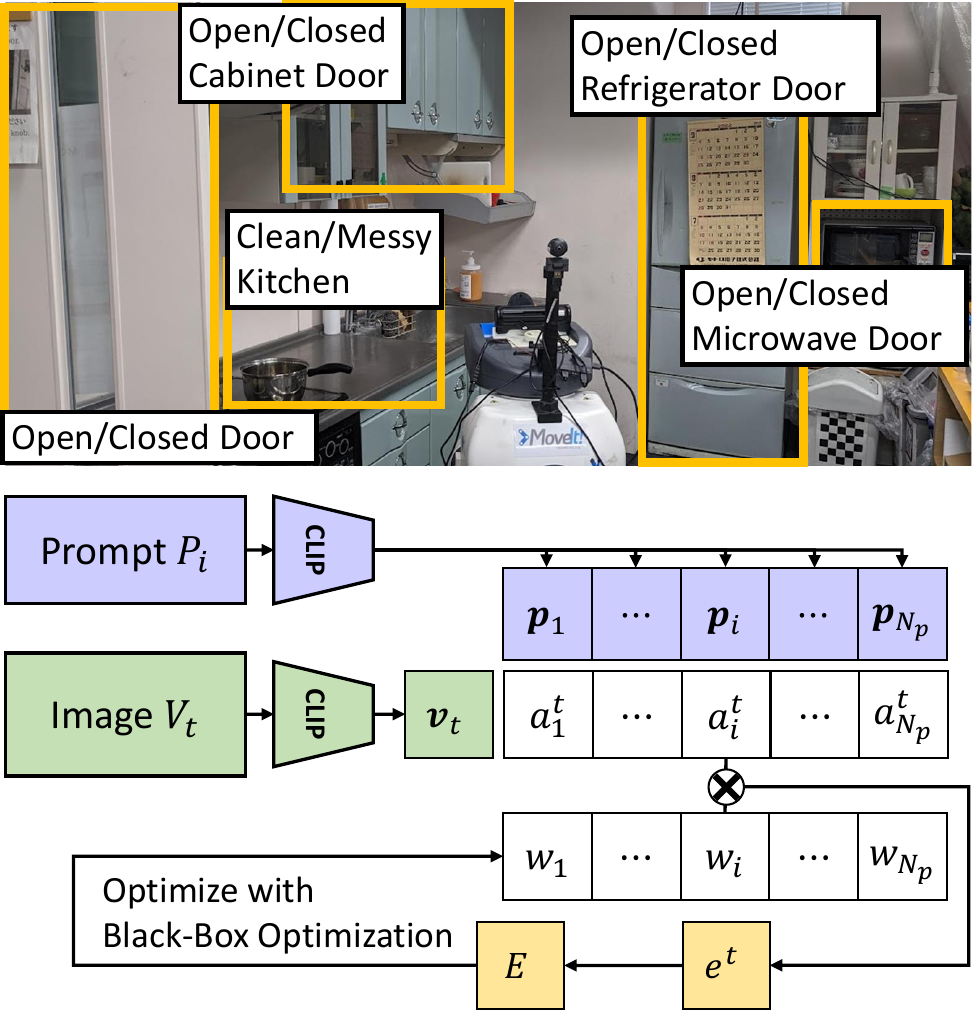}
  \vspace{-1.0ex}
  \caption{The concept of this study: for the robotic binary state recognition, we use the pre-trained vision-language model CLIP and black-box optimization to optimize the weights of prepared language prompts.}
  \vspace{-3.0ex}
  \label{figure:concept}
\end{figure}

\switchlanguage%
{%
  Therefore, we propose a method of state recognition using a large-scale vision-language model \cite{li2022largemodels, kawaharazuka2023ptvlm, kawaharazuka2024foundation} that is trained on a large visual and linguistic dataset and is capable of performing a wide variety of tasks without retraining, including Visual Question Answering (VQA) \cite{antol2015vqa, wang2022ofa}, Image-to-Text Retrieval (ITR) \cite{radford2021clip, girdhar2023imagebind}, etc.
  By using the spoken language for state recognition, even states that cannot be easily described by a program can be recognized.
  Although object recognition using vision-language models is common \cite{gu2022vild}, studies extending it to object and environment state recognition are lacking.
  Certainly, there are methods that implicitly perform state recognition within the trained reward and policy \cite{cui2022reward, xembodiment2024rtx}.
  However, our aim is to make this process explicit, enabling easy utilization in robot demonstrations.
  In this study, we perform state recognition utilizing CLIP \cite{radford2021clip}, which is a model capable of ITR.
  Note that VQA can also be utilized \cite{kawaharazuka2023ofaga}, but this takes too much time for inference, so we chose a model capable of ITR.
  By preparing multiple prompts that represent states of objects and environments in advance, and by calculating their similarity to the current image using CLIP, state recognition is possible (\figref{figure:concept}).
  Also, the recognition accuracy can be improved by extracting appropriate prompts from the prepared prompts \cite{allingham2023weighting}.
  Therefore, we prepare a small number of image datasets in advance and optimize the weight of each prompt by using black-box optimization.
  We also discuss the recognition accuracy according to the different evaluation functions in the black-box optimization.
  We show that this method is effective in recognizing the open/closed state of transparent doors, whether water is running or not, and even the qualitative state of whether the kitchen is clean or not, which have been challenging without manual programming, training individual neural networks, or having dedicated sensors.
  In addition, since only prompts and their weights need to be prepared for each recognizer, there is no need to prepare multiple models, which facilitates resource management.
  Note that this study does not focus on the concept of performing diverse state recognition simultaneously; instead, it emphasizes the concept of simplifying and intelligently executing each recognition component required for robot demonstrations.
  We believe that this research will revolutionize the recognition behavior of robots and facilitate the construction of more advanced and intelligent systems through the spoken language.
}%
{%
  そこで, 本研究では大規模視覚-言語モデル\cite{li2022largemodels, kawaharazuka2023ptvlm}に着目した.
  これは大規模な視覚と言語に関するデータセットから学習されており, Visual Question Answering \cite{antol2015vqa, wang2022ofa}やImage-to-Text Retrieval \cite{radford2021clip, girdhar2023imagebind}などを始めとした多様なタスクが再学習無しに実現可能である.
  視覚-言語モデルを用いた物体認識は一般的なタスクであるが, これを物体や環境の状態認識に拡張する研究は欠落している.
  もちろん, ロボットのrewardやpolicyの中でimplicitに状態認識を行う手法もあるが\cite{cui2022reward, xembodiment2024rtx}, これを明示的に行うことでロボットのデモンストレーションに容易に利用できるようにすることを目指す.
  本研究ではこの中でも, Image-to-Text Retrieval (ITR)が可能なモデルであるCLIP \cite{radford2021clip}を用いた二値状態認識を行う.
  事前に複数のプロンプトを用意し, 現在の画像についてCLIPにより類似度を計算, これを用いることで状態認識が可能となる(\figref{figure:concept}).
  一方で, 用意したプロンプト群から適切なプロンプトを抽出することで認識精度を向上させることが可能である.
  そのため, 予め認識したい状態の画像データセットを少数用意し, これと遺伝的アルゴリズムを用いることでそれぞれのプロンプトに関する重みを最適化する.
  遺伝的アルゴリズムにおける評価関数の違いに応じた認識精度についても議論を行う.
　本研究により, 手動のプログラミングやニューラルネットワークの再学習等をせずとも, 複数のプロンプトを用意するのみで二値状態認識が可能となることを示す.
  また, これまで個別のニューラルネットワークの学習や専用のセンサ等無しには難しかった蛇口からの出水状態や透明なドアの開閉状態, キッチンが綺麗かどうかといった質的状態の認識にも本研究が有効であることを示す.
  本研究はロボットの認識行動を大きく革新し, より高度で知能的なシステムの構築が容易になると確信する.
}%

\section{CLIP-based Robotic State Recognition Optimized with Black-Box Optimization} \label{sec:proposed}
\subsection{Pre-Trained Vision-Language Model - CLIP} \label{subsec:clip-model}
\switchlanguage%
{%
  We describe the basic usage of CLIP, a pre-trained vision-language model.
  CLIP is a model that calculates the similarity between an image and predefined texts (called prompts), and can search which prompt is closest to the current image.
  First, let $\bm{v}$ be a vector transformed by CLIP from the current image $V$, and let $\bm{p}_{\{1, 2, \cdots, N_{P}\}}$ be a vector transformed by CLIP from the set of predefined prompts $P_{\{1, 2, \cdots, N_{P}\}}$ ($N_{P}$ is the number of prompts).
  Next, the cosine similarity $a_{\{1, 2, \cdots, N_{P}\}}=\bm{v}^{T}\bm{p}_{\{1, 2, \cdots, N_{P}\}}$ between $\bm{v}$ and $\bm{p}$ is computed.
  Finally, we compute the softmax $\bm{s}_{i}$ for $a_{i}$ ($1 \leq i \leq N_{P}$) and take the most probable prompt $P_{i}$ as the text that best matches the current image $V$.
  This allows us to know, for example, whether the current image is more consistent with an apple or a banana by setting $P_{\{1, 2\}}=$\{apple, banana\}.
  The main concept of this study is to utilize CLIP for robotic state recognition through the spoken language.
}%
{%
  事前学習済み視覚-言語モデルであるCLIPの基本的な使い方について述べる.
  CLIPはある画像と予め定義された複数のテキスト(プロンプトと呼ばれる)の間の類似度を計算し, 現在の画像がどのプロンプトに最も近いかを検索可能なモデルである.
  まず, 現在の画像$V$をCLIPにより変換したベクトルを$\bm{v}$, 定義されたプロンプトの集合$P_{\{1, 2, \cdots, N_{P}\}}$をCLIPにより変換したベクトルを$\bm{p}_{\{1, 2, \cdots, N_{P}\}}$とする(ここで, $N_{P}$をプロンプト数とする).
  次に, これらの間のコサイン類似度$a_{\{1, 2, \cdots, N_{P}\}}=\bm{v}^{T}\bm{p}_{\{1, 2, \cdots, N_{P}\}}$を計算する.
  最後に, $a_{i}$ ($1 \leq i \leq N_{P}$)に関するソフトマックス$\bm{s}_{i}$を計算し, 最も確率の高かったプロンプト$P_{i}$を, 現在の画像$V$に最も合致するテキストとして取り出す.
  これにより, 例えば$P_{\{1, 2\}}=$\{リンゴ, バナナ\}とすることで, 現在の画像がリンゴとバナナ, どちらにより合致しているかを知ることができる.
  これを言語を介したロボットのための二値状態認識に利用するのが本研究の主なコンセプトである.
}%

\subsection{CLIP-based Robotic State Recognition} \label{subsec:state-recognition}
\switchlanguage%
{%
  For example, if you want to recognize the open/closed state of a door, you should set $P_{\{1, 2\}}=$\{``open door'', ``closed door''\}.
  Here, we can calculate the softmax $\bm{s}_{i}$ and judge that the door is open when $\bm{s}_{1}\geq\bm{s}_{2}$ and closed when $\bm{s}_{1}<\bm{s}_{2}$.
  This is the simplest method, but ``open door'' and ``closed door'' do not necessarily have the similar recognition performance.
  In that case, for example, $\bm{s}_{1}\geq\bm{s}_{2}$ will always hold regardless of whether the door is open or closed.
  Therefore, we consider another state recognition method.
  We prepare a set of prompts $P_{\{1, 2, \cdots, N_{P}\}}$, set the weighted sum $e=\Sigma_{i}w_{i}a_{i}$ of the obtained $a_{i}$ as the evaluation value ($w_{i}$ is the weight of the prompt $P_{i}$), and recognize that the door is open if it is above the threshold $C^{thre}$, and that the door is closed if it is below $C^{thre}$.
  $w_{i}$ is set to 1 for the prompt indicating that the door is open and -1 for the prompt indicating that the door is closed.
  Since there is no restriction on how to select prompts in this method, various prompt sets can be considered, and as long as appropriate prompts and threshold values are provided, state recognition is possible and performance can be easily adjusted.

  Here, we consider how to choose the prompt $P$.
  In this study, we consider changes in (i) articles, (ii) state expressions, (iii) wordings, and (iv) modifications to generate $P$.
  (i)  refers to the change in articles, e.g., for ``open door'': \{``an open door'', ``the open door'', ``this open door'', ``that open door''\}.
  (ii) refers to the change in the state expressions, e.g., for ``open door'': ``opening door'' and ``closed door''.
  (iii) refers to the change in wordings, e.g., for ``open door'': ``open entrance'' and ``open gate''.
  (iv) refers to the change in modifications, e.g., for ``open door'': ``open door of the room'' and ``open door of the house''.
  We can generate $P$ with various recognition performances using these changes.
}%
{%
  このCLIPを用いて二値状態認識を行う単純な方法を2種類: (i) and (ii)について述べる.
  (i) 例えば, ドアの開閉状態の認識を行いたい場合, $P_{\{1, 2\}}=$\{``open door'', ``closed door''\}とする.
  この際, 先のソフトマックス$\bm{s}_{i}$を計算し, $\bm{s}_{1}\geq\bm{s}_{2}$であるときドアは開いている, $\bm{s}_{1}<\bm{s}_{2}$のときドアは閉まっていると判断する方法である.
  これが最も単純な考え方であるが, それぞれのプロンプトは異なる認識性能を持っており, ``open door''と``closed door''が逆の似たような認識性能を持つとは限らない.
  その場合, ドアが開いていても閉じていても$\bm{s}_{1}\geq\bm{s}_{2}$または$\bm{s}_{1}<\bm{s}_{2}$の状態になってしまう.
  そのため, もう一つの状態認識方法を考える.
  (ii) あるプロンプト集合$P_{\{1, 2, \cdots, N_{P}\}}$を用意し, 得られた$a_{i}$の重み付け合計値$e=\Sigma_{i}w_{i}a_{i}$を評価値とし($w_{i}$はプロンプト$P_{i}$に対する重みとする), これが閾値$C^{thre}$以上であればドアが開いている, $C^{thre}$を下回ればドアが閉まっていると認識する方法である.
  $w_{i}$については, ドアが開いていることを示すプロンプトには1, ドアが閉まっていることを示すプロンプトには-1を与える.
  この方法であれば, プロンプトの選び方に制約がないため多様なプロンプト集合が考えられ, 適切なプロンプトと閾値さえ与えれば状態認識が可能となり性能を調整しやすい.

  ここでは, そのプロンプト$P$の選び方について考える.
  本研究では$P$の選び方として, (i)冠詞, (ii)状態表現,  (iii)単語, (iv)修飾を挙げる.
  (i)は, 例えば``open door''という単語を考えたとき, \{an open door, the open door, this open door, that open door\}等の冠詞変化を考えることを言う.
  (ii)は, 例えば``open door''の類義語``opening door''や対義語``closed door''等, その状態表現を変化させることを言う.
  (iii)は, 例えば``open door''の単語を変更したような, ``open entrance''や``open gate''などの別単語表現を用いることを言う.
  (iv)は, 例えば``open door''に修飾を加えたような, ``open door in the room''のような表現を用いることを言う.
  これらの方法により, 多様な認識性能を持つ$P$を生成することができる.
}%

  \begin{algorithm}[t]
    \caption{Calculation of $C^{thre}$}
    \label{algorithm:c-thre}
    \begin{algorithmic}[1]
      \Function{CalcCthre}{$e^{\{1, \cdots, T\}}, A^{\{1, \cdots, T\}}_{D}$}
        \State $b \gets 0$
        \State $b^{max} \gets 0$
        \State $C^{thre} \gets 0$
        \For{$i$ in $\textrm{argsort}^{des}(e^{\{1, \cdots, T\}})$}
          \State $b \gets b + A^{i}_{D}$
          \If{$b \geq b^{max}$}
            \State $b^{max} \gets b$
            \State $C^{thre} \gets (e^{i}+e^{i+1})/2$
          \EndIf
        \EndFor
        \State \Return $C^{thre}$
      \EndFunction
    \end{algorithmic}
  \end{algorithm}

\begin{figure*}[t]
  \centering
  \includegraphics[width=1.8\columnwidth]{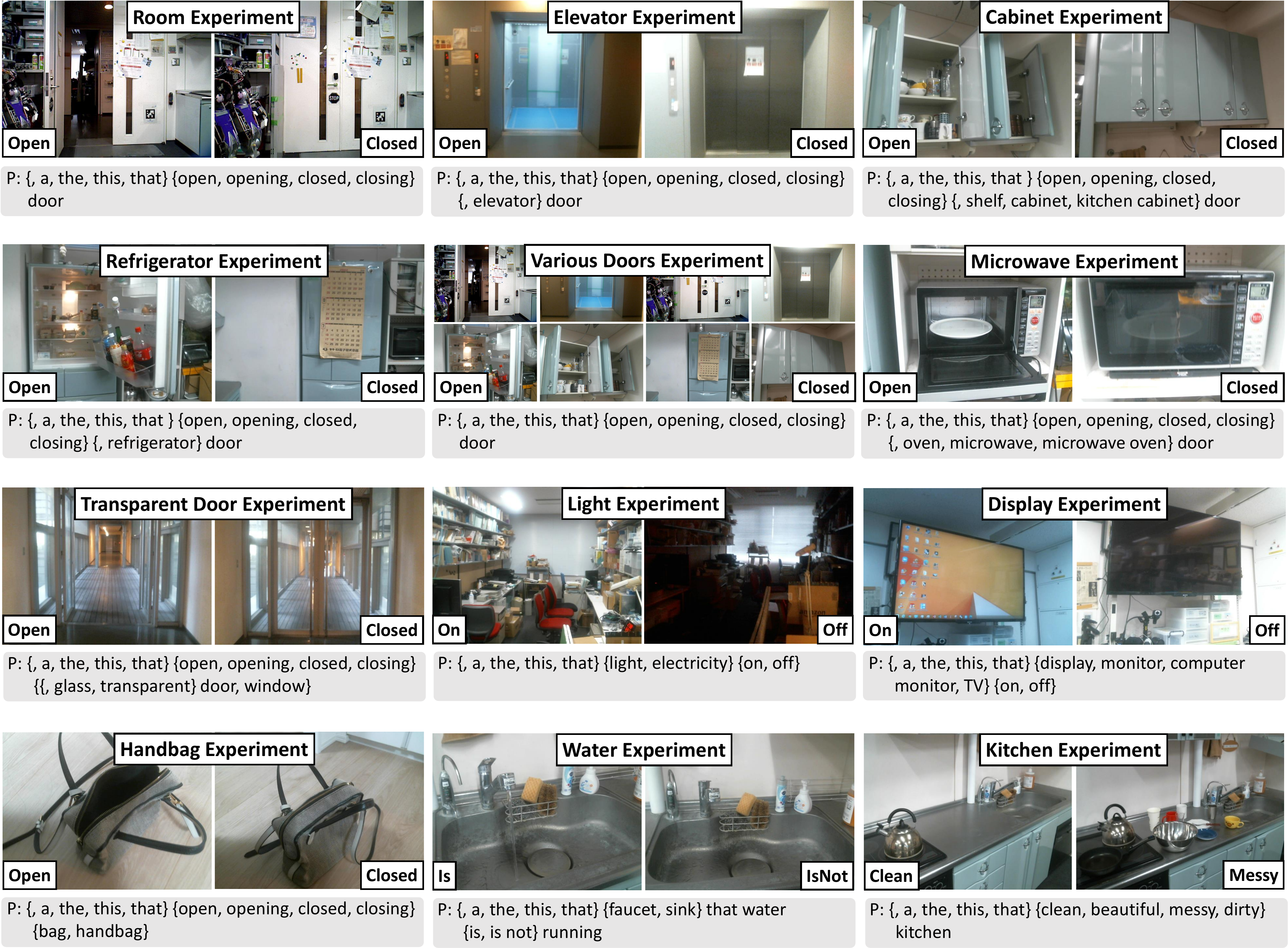}
  \vspace{-1.0ex}
  \caption{The set of prompts and representative images for the room, elevator, cabinet, refrigerator, various doors, microwave, transparent door, light, display, handbag, water, and kitchen experiments.}
  \vspace{-3.0ex}
  \label{figure:experiment}
\end{figure*}

\subsection{Optimization of CLIP-based Robotic State Recognition Using Black-Box Optimization} \label{subsec:bb-optimization}
\switchlanguage%
{%
  We describe a method to improve the recognition accuracy of state recognition using CLIP, based on the optimization of prompt weights by black-box optimization.
  For $e=\Sigma_{i}w_{i}a_{i}$ in the method described in \secref{subsec:state-recognition}, the weights $w_{i}$ were set manually to be appropriate.
  For example, if $P_{\{1, 2\}}=$\{``open door'', ``closed door''\}, then $w_{\{1, 2\}}=\{1, -1\}$.
  On the other hand, if $w_{i}$ is a continuous value and we can optimize this value, we should be able to construct a state recognizer with higher recognition accuracy.
  In this study, we optimize $w_{i}$ as a continuous value in the range [-1, 1].

  The optimization procedure is described below.
  First, as a dataset $D$, we prepare images $V_{t}$ ($1 \leq t \leq T$, where $t$ expresses the time axis, the angle of view, etc., and $T$ is the number of images), the correct response $A^{t}_{D}$ for each image $V_{t}$, and multiple prompts $P_{i}$ ($1 \leq i \leq N_{P}$).
  Here, $A^{t}_{D} \in \{1, -1\}$ (e.g., 1 for the open door state and -1 for the closed door state).

  Next, we define the evaluation function $E$ to be optimized.
  First, with the latent vector of the image $\bm{v}_{t}$, the latent vector of the prompt $\bm{p}_{i}$, and a given weight $w_{i}$, the most straightforward evaluation function $E_{1}$ that maximizes recognition accuracy when an optimal threshold $C^{thre}$ is set can be computed as follows,
  \begin{align}
    a^{t}_{i} &= \bm{v}^{T}_{t}\bm{p}_{i}\\
    e^{t} &= \Sigma_{i}w_{i}a^{t}_{i}/\Sigma_{i}|w_{i}|\\
    E_{1} &= \Sigma_{t}\textrm{bool}(A^{t}_{D}(e^{t}-C^{thre}) > 0) \label{eq:eval}
  \end{align}
  where $\textrm{bool}(\bullet)$ is a function that returns 1 when the condition $\bullet$ is satisfied and 0 otherwise.
  Also, $C^{thre}$ is a constant threshold value that can most accurately recognize the dataset $D$.
  The calculation method of $C^{thre}$ is shown in \algoref{algorithm:c-thre}.
  Here, $\textrm{argsort}^{des}(\bullet)$ is a function that returns the index of $\bullet$ when sorted in descending order.
  That is, $e^{t}$ is sorted with respect to $t$, and $C^{thre}$ is sequentially set as the value between two consecutive $e^{t}$, and $C^{thre}$ with the largest $E_{1}$ is searched.
  Second, the evaluation function $E_{2}$ that maximizes not only recognition accuracy but also the difference of the means $\mu_{\{-1, 1\}}$ of $e^{t}$ of data for which $A^{t}_{D}$ is 1 or -1 is calculated as follows,
  \begin{align}
    \mu_{1} &= \Sigma_{t}(\textrm{bool}(A^{t}_{D}\geq0)(e^{t}-C^{thre}))\\
    \mu_{-1} &= \Sigma_{t}(\textrm{bool}(A^{t}_{D}<0)(e^{t}-C^{thre}))\\
    \mu &= \mu_{1}-\mu_{-1}\\
    E_{2} &= E_{1} + \alpha_{2}\mu
  \end{align}
  where $\alpha_{2}$ is a constant value, which is set as $\alpha_{2}=1.0$ in this study.
  Third, in addition to maximizing the difference of means, we compute the evaluation function $E_{3}$ that minimizes the standard deviations $\sigma_{\{-1, 1\}}$ of $e^{t}$ of data for which $A^{t}_{D}$ is 1 or -1 as follows,
  \begin{align}
    \sigma_{1} &= \textrm{std}_{t}(\textrm{bool}(A^{t}_{D}\geq1)(e^{t}-C^{thre}))\\
    \sigma_{-1} &= \textrm{std}_{t}(\textrm{bool}(A^{t}_{D}<-1)(e^{t}-C^{thre}))\\
    \sigma &= \sigma_{1}\sigma_{-1}\\
    E_{3} &= E_{1} + \alpha_{3}\mu/\sigma
  \end{align}
  where $\textrm{std}_{t}(\bullet)$ is a function to calculate the standard deviation of $\bullet$ regarding $t$, and $\alpha_{3}$ is a constant value, which is set as $\alpha_{3}=0.00001$ in this study.
  We compare these three types of evaluation functions $E_{\{1, 2, 3\}}$.

  Finally, we optimize the weight $w_{i}$ by a genetic algorithm, one of the black-box optimization methods that maximize the evaluation function $E$.
  We use Deap \cite{fortin2012deap} as a library and perform crossover with a probability of 50\% by the function cxBlend and mutation with a probability of 20\% by the function mutGaussian with mean 0 and standard deviation 0.5.
  Individuals are selected by the function selTournament, and the tournament size is set to 3.
  The number of individuals and generations are each set to 300.
}%
{%
  CLIPを用いた二値状態認識について, 遺伝的アルゴリズムによるプロンプトの重み付け最適化に基づく, 認識精度の向上方法について述べる.
  \secref{subsec:state-recognition}で述べた方法では, $e=\Sigma_{i}w_{i}a_{i}$について, 重み$w_{i}$は人間が適切な重みを設定していた.
  例えばドアの開閉状態認識について, $P_{\{1, 2\}}=$\{``open door'', ``closed door''\}と設定した場合, 一般的には$w_{\{1, 2\}}=\{1, -1\}$である.
  一方で, もし$w_{i}$が連続値であり, この値を最適化することができれば, より認識精度の高い状態認識器を構成できるはずである.
  本研究では, $w_{i}$を[-1, 1]の範囲の連続値として最適化を行う.

  最適化の手順を以下に述べる.
  まず, データセット$D$として画像$V_{t}$ ($1 \leq t \leq T$. $t$は時間軸や画像を撮る姿勢等を表し, $T$は画像の枚数である)と, それぞれの画像について正しい回答$A^{t}_{D}$, 複数のプロンプト$P_{i}$ ($1 \leq i \leq N_{P}$)を用意する.
  ここで, $A^{t}_{D}$は\{1, -1\}とする(ドアの開閉であれば, 例えば1を開いた状態, -1を閉じた状態とする).

  次に, 最適化する評価関数$E$を定める.
  まず, 画像データセットの潜在空間$\bm{v}_{t}$とプロンプトの潜在空間$\bm{p}_{i}$, ある重み$w_{i}$が用意された状態で, 最適な閾値$C^{thre}$を設定した際の認識性能を最大化する最も単純な評価関数$E_{1}$は以下のように計算できる.
  \begin{align}
    a^{t}_{i} &= \bm{v}^{T}_{t}\bm{p}_{i}\\
    e^{t} &= \Sigma_{i}w_{i}a^{t}_{i}/\Sigma_{i}|w_{i}|\\
    E_{1} &= \Sigma_{t}\textrm{bool}(A^{t}_{D}(e^{t}-C^{thre}) > 0) \label{eq:eval}
  \end{align}
  ここで, $\textrm{bool}(\bullet)$は, 条件$\bullet$を満たすとき1, それ以外のとき0を返す関数である.
  また, $C^{thre}$はある定数であり, この$C^{thre}$はデータセット$D$を最も正確に認識できる閾値とする.
  $C^{thre}$の計算方法を\algoref{algorithm:c-thre}に示す.
  ここで, $\textrm{argsort}^{des}(\bullet)$は$\bullet$を降順にソートした際のインデックスを返す関数である.
  つまり, $e^{t}$を$t$についてソートし, 連続する2つの値の中間の値を順に$C^{thre}$として, $E_{1}$が最も大きくなるような$C^{thre}$を探索している.
  次に, $A^{t}_{D}$が1または-1であるデータの$e^{t}$の平均$\mu$の差を最大化する評価関数$E_{2}$を以下のように計算する.
  \begin{align}
    \mu_{1} &= \Sigma_{t}(\textrm{bool}(A^{t}_{D}\geq0)(e^{t}-C^{thre}))\\
    \mu_{-1} &= \Sigma_{t}(\textrm{bool}(A^{t}_{D}<0)(e^{t}-C^{thre}))\\
    \mu &= \mu_{1}-\mu_{-1}\\
    E_{2} &= E + \alpha_{2}\mu
  \end{align}
  ここで, $\alpha_{2}$はある定数であり, 本研究では$\alpha_{2}=1.0$とした.
  最後に, 平均の差の最大化に加え, $A^{t}_{D}$が1または-1であるそれぞれのデータ集合における$e^{t}$の標準偏差$\sigma$を最小化する評価関数$E_{3}$を以下のように計算する.
  \begin{align}
    \sigma_{1} &= \textrm{std}_{t}(\textrm{bool}(A^{t}_{D}\geq1)(e^{t}-C^{thre}))\\
    \sigma_{-1} &= \textrm{std}_{t}(\textrm{bool}(A^{t}_{D}<-1)(e^{t}-C^{thre}))\\
    \sigma &= \sigma_{1}\sigma_{-1}\\
    E_{3} &= E + \alpha_{3}\mu/\sigma
  \end{align}
  ここで, $\textrm{std}_{t}(\bullet)$は$t$について$\bullet$の標準偏差を計算する関数, $\alpha_{3}$はある定数であり, 本研究では$\alpha_{3}=0.00001$とした.
  本研究では, この3種類の評価関数$E_{\{1, 2, 3\}}$についての比較も行う.

  最後に, この評価関数$E$を最大化する遺伝的アルゴリズムにより重み$w_{i}$を最適化する.
  ライブラリとしてDeap \cite{fortin2012deap}を用い, 50\%の確率で関数cxBlendにより交叉, 20\%の確率で平均0, 標準偏差0.5の関数mutGaussianにより突然変異を行う.
  個体選択は関数selTournamentで行い, トーナメントサイズを3とする.
  また, 個体数は300, 世代数は300とする.
}%

\section{Experiment} \label{subsec:experiments}
\switchlanguage%
{%
  The representative images used in our state recognition experiments and their prompt combinations are shown in \figref{figure:experiment}.
  Specifically, we conduct experiments to recognize whether a room door, an elevator door, a cabinet door, a refrigerator door, various doors, a microwave door, and a transparent door are open or closed, whether a light and a display are on or off, whether a handbag is open or closed, whether water is running or not from a faucet, and whether a kitchen is clean or not.
  As a dataset $D_{opt}$ for optimization, we have prepared a total of 20 images, e.g., 10 each of the open and closed states of the door.
  For ``Various Doors Experiment'', 80 pictures are prepared, 20 pictures for each of the above four doors: room, elevator, cabinet, and refrigerator doors.
  We also prepared a dataset $D_{eval}$ for evaluation, which is different from $D_{opt}$ but includes the same number of images.

  In this study, we mainly compare the results of recognition experiments using $w_{i}$ optimized by black-box optimization based on the evaluation functions $E_{\{1, 2, 3\}}$ (denoted as OPT-\{1, 2, 3\}).
  In addition, we also present results for the case where all the prepared $P$ are used equally ($w_{i}=\{-1, 1\}$, the same with \secref{subsec:state-recognition}, denoted as ALL), and for the case where only the best $P$ that maximizes $E$ is used (denoted as ONE).
  We denote the rate of correct state recognitions in each dataset $D_{\{opt, eval\}}$ as $R_{\{opt, eval\}}$, and compare the results for each method OPT-1, OPT-2, OPT-3, ALL, and ONE.
  Note that in many cases, the state recognition problem discussed in this paper can be performed with an accuracy of close to 100\%, if human annotation, point cloud processing, and dedicated sensors are skillfully utilized.
  On the other hand, an important point of this study is that various state recognitions can be easily achieved without re-training neural networks or manually programming through the spoken language, and only prompts and their weights need to be prepared for each recognizer, thus eliminating the need to prepare multiple models and facilitating resource management.
}%
{%
  本研究の二値状態認識実験で用いた画像の一部とそのプロンプトの組み合わせを\figref{figure:experiment}に示す.
  具体的には, 標準的なドア・エレベータのドア・棚のドア・冷蔵庫のドア・様々なドア・レンジのドア・透明なドアの開閉, 明かり・ディスプレイのオンオフ, カバンの開閉, 蛇口からの出水, キッチンの綺麗さを認識する実験を行う.
  最適化用データセット$D_{opt}$として, 例えばドアについては開いている状態と閉じている状態を10枚ずつの計20枚, 水については蛇口から出ている状態と出ていない状態を10枚ずつの計20枚を用意した.
  また, 評価用データセット$D_{eval}$として, $D_{opt}$とは異なる同じ枚数のデータを用意した.

  本研究では主に, $E_{\{1, 2, 3\}}$の評価関数に基づき遺伝的アルゴリズムにより最適化された$w_{i}$を用いた際の認識実験結果OPT-\{1, 2, 3\}を比較する.
  それに加え, 用意した$P$を全て均等に使う場合($w_{i}=\{-1, 1\}$. \secref{subsec:state-recognition}と同様. ALLと呼ぶ), 用意した$P$の中から$E$を最大化する最善の$P$を一つだけ使う場合(ONEと呼ぶ)の結果も述べる.
  それぞれのデータセット$D_{\{opt, eval\}}$における二値状態認識の正解率を$R_{\{opt, eval\}}$と表し, これをそれぞれの手法OPT-1, OPT-2, OPT-3, ALL, ONEについて比較する.
  なお, ここで扱う状態認識は人間によるアノテーションや点群処理, 専用のセンサを巧みに駆使すれば, 方法や設定次第では100\%近い高い精度の認識が可能な場合も多い.
  一方, 本研究の重要な点は, 言語を介することでニューラルネットワークの再学習や人手によるプログラミング等無しに多様な状態認識が容易に可能な点である.
}%

\subsection{Room Experiment}
\switchlanguage%
{%
  The results of the open/closed state recognition experiment of a room door are shown in \tabref{table:door}.
  The recognition accuracies of OPT-\{1, 2, 3\} are all similarly high, and $R_{opt} > R_{eval}$ holds.
  Also, OPT $>$ ONE $>$ ALL holds for the recognition accuracy, and the accuracy without optimization is lower than that with optimization.
}%
{%
  通常のドア開閉の状態認識実験の結果を\tabref{table:door}に示す.
  OPT-\{1, 2, 3\}の認識精度は全て同様に高く, $R_{opt} > R_{eval}$が成り立つ.
  また, 認識精度はOPT $>$ ONE $>$ ALLが成り立ち, 最適化を行わない場合は行う場合に比べて精度が低い.
}%


\begin{table}[htb]
  \centering
  \caption{The result of the room experiment}
  \begin{tabular}{l||c|c|c|c|c}
    & OPT-1 & OPT-2 & OPT-3 & ALL & ONE \\ \hline\hline
    $R_{opt}$ [\%]  & 100 & 100 & 100 & 85 & 95\\\hline
    $R_{eval}$ [\%] &  90 &  90 &  90 & 80 & 85\\
  \end{tabular}
  \label{table:door}
\end{table}

\subsection{Elevator Experiment}
\switchlanguage%
{%
  The results of the open/closed state recognition experiment of an elevator door are shown in \tabref{table:elevator}.
  The recognition accuracies of OPT-\{1, 2, 3\}, ONE, and ALL are almost the same, and $R_{opt} \simeq R_{eval}$ holds.
  Even without optimization, state recognition can be performed with high accuracy.
}%
{%
  エレベータのドア開閉の状態認識実験の結果を\tabref{table:elevator}に示す.
  OPT-\{1, 2, 3\}, ONE, ALLの認識精度はほぼ全て同様であり, $R_{opt} \simeq R_{eval}$が成り立つ.
  最適化を行わない場合でも高い精度で状態認識が可能であった.
}%


\begin{table}[htb]
  \centering
  \caption{The result of the elevator experiment}
  \begin{tabular}{l||c|c|c|c|c}
    & OPT-1 & OPT-2 & OPT-3 & ALL & ONE \\ \hline\hline
    $R_{opt}$ [\%]  & 100 & 100 & 100 & 100 & 100\\\hline
    $R_{eval}$ [\%] & 100 &  95 & 100 & 100 &  95\\
  \end{tabular}
  \label{table:elevator}
\end{table}

\subsection{Cabinet Experiment}
\switchlanguage%
{%
  The results of the open/closed state recognition experiment of a cabinet door are shown in \tabref{table:cabinet}.
  The accuracy trend is similar to that of \tabref{table:door}.
  The recognition accuracies of OPT-\{1, 2, 3\} are similarly high, and $R_{opt} > R_{eval}$ holds.
  Also, OPT $\simeq$ ONE $>$ ALL holds for the recognition accuracy.
}%
{%
  棚のドア開閉の状態認識実験の結果を\tabref{table:cabinet}に示す.
  その精度は\tabref{table:door}と似通っている.
  OPT-\{1, 2, 3\}の認識精度はほぼ全て同様に高く, $R_{opt} > R_{eval}$が成り立つ.
  また, 認識精度はOPT $\simeq$ ONE $>$ ALLが成り立つ.
}%


\begin{table}[htb]
  \centering
  \caption{The result of the cabinet experiment}
  \begin{tabular}{l||c|c|c|c|c}
    & OPT-1 & OPT-2 & OPT-3 & ALL & ONE \\ \hline\hline
    $R_{opt}$ [\%]  & 100 & 100 & 100 & 85 & 100\\\hline
    $R_{eval}$ [\%] &  85 &  90 &  85 & 80 &  85\\
  \end{tabular}
  \label{table:cabinet}
\end{table}

\subsection{Refrigerator Experiment}
\switchlanguage%
{%
  The results of the open/closed state recognition experiment of a refrigerator door are shown in \tabref{table:refrigerator}.
  The recognition accuracies of OPT-\{1, 2, 3\} are similarly high, and $R_{opt} \simeq R_{eval}$ holds.
  On the other hand, unlike the previous cases, the recognition accuracy of ALL is higher than that of ONE.
  Moreover, $R_{opt} > R_{eval}$ holds for ALL and ONE.
}%
{%
  冷蔵庫のドア開閉の状態認識実験の結果を\tabref{table:refrigerator}に示す.
  OPT-\{1, 2, 3\}については認識精度はほぼ全て同様に高く, $R_{opt} \simeq R_{eval}$が成り立つ.
  一方これまでとは異なり, ALLの方がONEよりも認識精度は高い.
  また, ALLとONEについては$R_{opt} > R_{eval}$が成り立つ.
}%


\begin{table}[htb]
  \centering
  \caption{The result of the refrigerator experiment}
  \begin{tabular}{l||c|c|c|c|c}
    & OPT-1 & OPT-2 & OPT-3 & ALL & ONE \\ \hline\hline
    $R_{opt}$ [\%]  & 100 & 100 & 100 & 100 & 90\\\hline
    $R_{eval}$ [\%] & 100 &  95 & 100 &  85 & 80\\
  \end{tabular}
  \label{table:refrigerator}
\end{table}

\subsection{Various Doors Experiment}
\switchlanguage%
{%
  The results of the open/closed state recognition experiment of the aforementioned four types of doors using the same prompts are shown in \tabref{table:various}.
  For OPT-\{1, 2, 3\}, the states of four different doors are recognized with high accuracy even with the same prompts.
  Also, $R_{opt} > R_{eval}$ holds, among which OPT-3 $>$ OPT-2 $>$ OPT-1 holds for $R_{eval}$.
  On the other hand, the recognition accuracies of ONE and ALL are much lower than those of OPT, and ALL $>$ ONE holds for the recognition accuracy.
}%
{%
  前述の通常のドア, エレベータのドア, 棚のドア, 冷蔵庫のドア全ての開閉の状態認識実験を同じプロンプトを用いて行った結果を\tabref{table:various}に示す.
  OPT-\{1, 2, 3\}については, 全く同じプロンプトでも4種類の異なるドアの開閉状態が高い精度で認識できている.
  また, $R_{opt} > R_{eval}$が成り立つが, その中でも$R_{eval}$についてはOPT-3 $>$ OPT-2 $>$ OPT-1が成り立つ.
  一方で, ONEとALLの認識精度はOPTに比べると大きく劣っており, 認識精度はALL $>$ ONEであった.
}%


\begin{table}[htb]
  \centering
  \caption{The result of the various doors experiment}
  \begin{tabular}{l||c|c|c|c|c}
    & OPT-1 & OPT-2 & OPT-3 & ALL & ONE \\ \hline\hline
    $R_{opt}$ [\%]  & 99 & 99 & 99 & 84 & 70\\\hline
    $R_{eval}$ [\%] & 91 & 93 & 96 & 74 & 69\\
  \end{tabular}
  \label{table:various}
\end{table}

\subsection{Microwave Experiment}
\switchlanguage%
{%
  The results of the open/closed state recognition experiment of a microwave door are shown in \tabref{table:microwave}.
  OPT-2 $>$ OPT-3 $>$ OPT-1 holds for the recognition accuracy, and especially for OPT-3, $R_{eval}$ is much lower than $R_{opt}$.
  Also, while the accuracy of ONE is close to that of OPT, the accuracy of ALL is significantly lower.
}%
{%
  電子レンジのドア開閉の状態認識実験の結果を\tabref{table:microwave}に示す.
  認識精度はOPT-2 $>$ OPT-3 $>$ OPT-1の順で高いが, 特にOPT-3に関しては$R_{eval}$が$R_{opt}$に比べて大きく落ちている.
  また, ONEはOPTに近い精度な一方, ALLの精度が著しく低い.
}%


\begin{table}[htb]
  \centering
  \caption{The result of the microwave experiment}
  \begin{tabular}{l||c|c|c|c|c}
    & OPT-1 & OPT-2 & OPT-3 & ALL & ONE \\ \hline\hline
    $R_{opt}$ [\%]  & 90 & 100 & 100 & 60 & 95\\\hline
    $R_{eval}$ [\%] & 75 &  90 &  75 & 50 & 80\\
  \end{tabular}
  \label{table:microwave}
\end{table}

\subsection{Transparent Door Experiment}
\switchlanguage%
{%
  The results of the open/closed state recognition experiment of a transparent glass door are shown in \tabref{table:transparent}.
  This task is more difficult than the previous ones, and $R_{opt}$ does not reach its maximum value even after optimization.
  OPT-2 $>$ OPT-3 $>$ OPT-1 holds for the recognition accuracy, and $R_{opt} > R_{eval}$ holds for OPT-1 and OPT-3, while $R_{opt} = R_{eval}$ for OPT-2, maintaining high accuracy.
  As with \tabref{table:microwave}, the accuracy of ONE is close to that of OPT, while the accuracy of ALL is significantly lower.
}%
{%
  透明なガラスドア開閉の状態認識実験の結果を\tabref{table:transparent}に示す.
  これまでのタスクに比べると難しく, 最適化を行った場合にも$R_{opt}$が最大値に達していない.
  認識精度はOPT-2 $>$ OPT-3 $>$ OPT-1の順で高く, OPT-1とOPT-3については$R_{opt} > R_{eval}$なのに対して, OPT-2については$R_{opt} = R_{eval}$であり高い精度を保っている.
  また, \tabref{table:microwave}同様ONEはOPTに近い精度な一方, ALLの精度が著しく低い.
}%


\begin{table}[htb]
  \centering
  \caption{The result of the transparent door experiment}
  \begin{tabular}{l||c|c|c|c|c}
    & OPT-1 & OPT-2 & OPT-3 & ALL & ONE \\ \hline\hline
    $R_{opt}$ [\%]  & 85 & 90 & 90 & 55 & 90\\\hline
    $R_{eval}$ [\%] & 70 & 90 & 75 & 50 & 80\\
  \end{tabular}
  \label{table:transparent}
\end{table}

\subsection{Light Experiment}
\switchlanguage%
{%
  In the following sections, we show the results of state recognition experiments that are qualitatively different from the open/closed door state recognition experiments.
  The results of the on/off state recognition experiment of a light are shown in \tabref{table:light}.
  While the recognition accuracies of OPT-\{1, 2, 3\} and ONE are similarly high, that of ALL is lower.
  Also, $R_{opt} > R_{eval}$ holds.
}%
{%
  以降はこれまでのドアの開閉状態認識実験とは質的に異なる状態認識の結果について示す.
  電気のオンオフ状態認識実験の結果を\tabref{table:light}に示す.
  OPT-\{1, 2, 3\}, ONEの認識精度はほぼ全て同様に高い一方, ALLの精度はそれらと比べると低い.
  また, 全体を通して$R_{opt} > R_{eval}$が成り立つ.
}%


\begin{table}[htb]
  \centering
  \caption{The result of the light experiment}
  \begin{tabular}{l||c|c|c|c|c}
    & OPT-1 & OPT-2 & OPT-3 & ALL & ONE \\ \hline\hline
    $R_{opt}$ [\%]  & 100 & 100 & 100 & 75 & 100\\\hline
    $R_{eval}$ [\%] &  95 &  95 & 100 & 80 &  95\\
  \end{tabular}
  \label{table:light}
\end{table}

\subsection{Display Experiment}
\switchlanguage%
{%
  The results of the on/off state recognition experiment of a display are shown in \tabref{table:display}.
  The recognition accuracies of OPT-\{1, 2, 3\}, ONE, and ALL are similarly high.
  Also, $R_{opt} > R_{eval}$ holds.
}%
{%
  ディスプレイのオンオフ状態認識実験の結果を\tabref{table:display}に示す.
  OPT-\{1, 2, 3\}, ONE, ALLの認識精度はほぼ全て同様であり, 高い精度で認識に成功している.
  また, 全体を通して$R_{opt} > R_{eval}$が成り立つ.
}%


\begin{table}[htb]
  \centering
  \caption{The result of the display experiment}
  \begin{tabular}{l||c|c|c|c|c}
    & OPT-1 & OPT-2 & OPT-3 & ALL & ONE \\ \hline\hline
    $R_{opt}$ [\%]  & 100 & 100 & 100 & 100 & 100\\\hline
    $R_{eval}$ [\%] &  95 &  95 & 100 &  95 &  90\\
  \end{tabular}
  \label{table:display}
\end{table}

\subsection{Handbag Experiment}
\switchlanguage%
{%
  The results of the open/closed state recognition experiment of a handbag are shown in \tabref{table:handbag}.
  The recognition accuracies of OPT-\{1, 2, 3\} are similarly high with respect to $R_{opt}$, but $R_{eval}$ is low for OPT-3.
  OPT $>$ ALL $>$ ONE holds for the recognition accuracy, and especially $R_{eval}$ is low for ONE.
}%
{%
  ハンドバッグのチャック開閉の状態認識実験の結果を\tabref{table:handbag}に示す.
  OPT-\{1, 2, 3\}の認識精度は$R_{opt}$に関して全て同様に高いが, OPT-3のみ$R_{eval}$が低い.
  認識精度はOPT $>$ ALL $>$ ONEであり, 特にONEは$R_{eval}$が低い.
}%


\begin{table}[htb]
  \centering
  \caption{The result of the handbag experiment}
  \begin{tabular}{l||c|c|c|c|c}
    & OPT-1 & OPT-2 & OPT-3 & ALL & ONE \\ \hline\hline
    $R_{opt}$ [\%]  & 100 & 100 & 100 & 85 & 85\\\hline
    $R_{eval}$ [\%] &  90 &  90 &  80 & 80 & 65\\
  \end{tabular}
  \label{table:handbag}
\end{table}

\subsection{Water Experiment}
\switchlanguage%
{%
  The results of the state recognition experiment of whether water is running or not are shown in \tabref{table:water}.
  The recognition accuracies of $R_{opt}$ for OPT-\{1, 2, 3\} are similarly high, while OPT-3 $>$ OPT-2 $>$ OPT-1 holds for $R_{eval}$.
  OPT $>$ ONE $>$ ALL holds for the recognition accuracy, and especially $R_{opt}$ of ALL is low.
  Also, $R_{eval}$ is much lower than $R_{opt}$, and the adaptability to the dataset for evaluation is low.
}%
{%
  蛇口からの出水状態認識実験の結果を\tabref{table:water}に示す.
  OPT-\{1, 2, 3\}の認識精度は$R_{opt}$については全て同様に高いが, $R_{eval}$についてはOPT-3 $>$ OPT-2 $>$ OPT-1が成り立つ.
  認識精度はOPT $>$ ONE $>$ ALLが成り立ち, 特にALLの$R_{opt}$は低い.
  また, 全体を通して$R_{opt}$に比べて$R_{eval}$が大きく下がっており, 評価用データセットへの適応性が低い.
}%


\begin{table}[htb]
  \centering
  \caption{The result of the water experiment}
  \begin{tabular}{l||c|c|c|c|c}
    & OPT-1 & OPT-2 & OPT-3 & ALL & ONE \\ \hline\hline
    $R_{opt}$ [\%]  & 100 & 100 & 100 & 55 & 90\\\hline
    $R_{eval}$ [\%] &  65 &  70 &  75 & 70 & 70\\
  \end{tabular}
  \label{table:water}
\end{table}

\subsection{Kitchen Experiment}
\switchlanguage%
{%
  The results of the state recognition experiment of whether a kitchen is clean or not are shown in \tabref{table:kitchen}.
  OPT-2 $>$ OPT-3 $>$ OPT-1 holds for the recognition accuracy, and perfect recognition is achieved for OPT-2.
  The performance of ALL and ONE is worse than that of OPT.
}%
{%
  キッチンの整理整頓状態認識実験の結果を\tabref{table:kitchen}に示す.
  OPT-2 $>$ OPT-3 $>$ OPT-1の順に認識精度が高く, OPT-2については完璧な認識ができている.
  ALLとONEについてはOPTよりも性能が落ちている.
}%


\begin{table}[htb]
  \centering
  \caption{The result of the kitchen experiment}
  \begin{tabular}{l||c|c|c|c|c}
    & OPT-1 & OPT-2 & OPT-3 & ALL & ONE \\ \hline\hline
    $R_{opt}$ [\%]  & 90 & 100 & 100 & 80 & 85\\\hline
    $R_{eval}$ [\%] & 80 & 100 &  85 & 85 & 70\\
  \end{tabular}
  \label{table:kitchen}
\end{table}

\section{Discussion} \label{sec:discussion}
\begin{figure}[t]
  \centering
  \includegraphics[width=1.0\columnwidth]{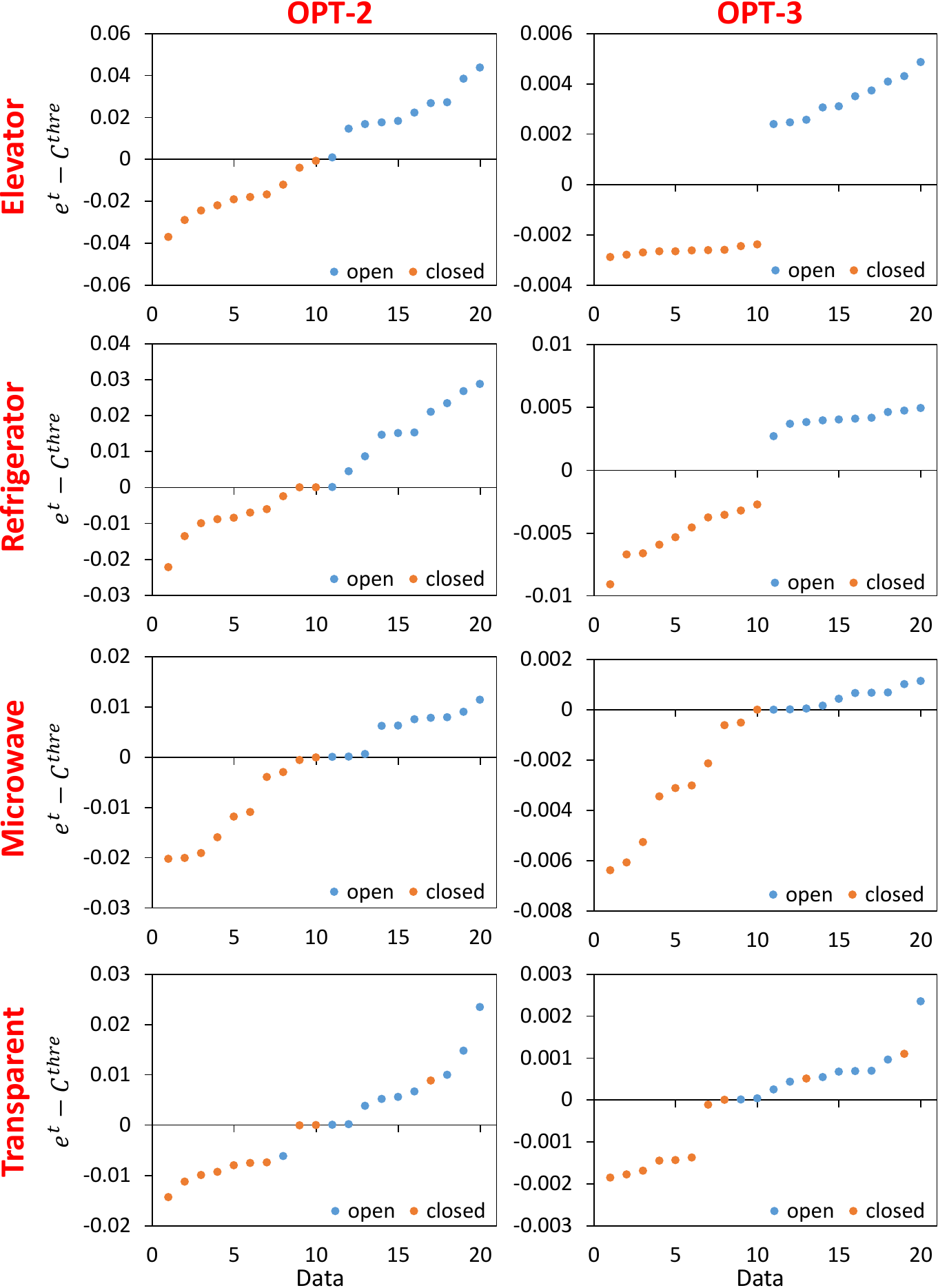}
  \vspace{-1.0ex}
  \caption{The relationship between $e^{t}-C^{thre}$ and open/closed labels $A^{t}_{D}$ of the dataset for optimization $D_{opt}$.}
  \vspace{-3.0ex}
  \label{figure:discussion}
\end{figure}

\switchlanguage%
{%
  We were able to obtain highly accurate state recognizers throughout the experiments, and several characteristics were observed.
  First, in most cases, OPT has better recognition performance than ALL using all prompts or ONE with only one best prompt, indicating the importance of weighting optimization of prompts.
  In OPT, the same prompt can be used to recognize a variety of open/closed door states, and can recognize even the state of a transparent door or the qualitative clean/messy state of a kitchen.
  The performance difference between ALL and ONE varies from task to task and cannot be judged in general.
  On the other hand, the performance of ALL and ONE is high for simple tasks, and in particular, ALL is the easiest to use because it does not require weighting decisions.

  Next, for $R_{opt}$, the recognition accuracy of most tasks is 100\% in the case of OPT.
  On the other hand, there are some differences in the performance for $R_{eval}$, and we found three trends (1)--(3).
  (1) OPT-1 does not outperform OPT-2 and OPT-3 by itself.
  (2) The recognition accuracy of OPT-3 is higher than that of OPT-2 for tasks with generally high recognition rate, i.e. the refrigerator, various doors, light, and display experiments (except for in the water experiment), although the difference is not large.
  (3) The recognition accuracy of OPT-2 is higher than that of OPT-3 for tasks with generally low recognition rates, i.e. the cabinet, microwave, transparent door, handbag, kitchen experiment.
  Let us discuss the trends (2) and (3) using \figref{figure:discussion}.
  \figref{figure:discussion} shows the label $A^{t}_{D}$ of the open/closed states and $e^{t}-C^{thre}$ for 20 images of $D_{opt}$, sorted by $e^{t}-C^{thre}$.
  The data above the line $e^{t}-C^{thre}=0$ is recognized as open and below as closed.
  Here, the upper two graphs in \figref{figure:discussion} correspond to (2), and the lower two graphs to (3).
  In the elevator and refrigerator experiments, it can be seen that the boundary is clearer in OPT-3 than in OPT-2 by considering the minimization of variance.
  This makes the recognizer robust, and OPT-3 outperforms OPT-2 in $R_{eval}$.
  On the other hand, in the microwave and transparent door experiments, a clear boundary cannot be created either due to the failure of the variance minimization or due to the inclusion of values that cannot be correctly classified.
  In this case, OPT-2, which prioritizes the maximization of the difference of means, performs better than OPT-3.
  Note that since the difference of performance in (2) is not as large as that in (3), the performance will be most stable when OPT-2 is used.

  We discuss the limitations of this study.
  We found that OPT-2 or OPT-3 can achieve high performance in state recognition for most of the tasks in this study.
  On the other hand, there is still room for improvement, especially in water recognition, and $R_{eval}$ should be improved for all tasks.
  While it is possible that better performance could be achieved by using tasks like Visual Question Answering (VQA) for these, there is also the issue of the time required for inference \cite{kawaharazuka2023ofaga}.
  We believe that the performance of this method will be greatly improved by the future development of large-scale vision-language models.
  Also, it is necessary to develop a more practical system using this method in the future.
  As one example, we conducted a simple experiment in a supplementary video.
  We have performed a patrol experiment using the proposed state recognition on the mobile robot Fetch.
  The robot can close the refrigerator door if it is open, close the cabinet door if it is open, and exit the room if the door is open.
  Note that the robot position of executing the state recognition and the robot motion of closing the door are prepared in advance.
  If the method can handle not only the state of the environment and objects but also the states of the robot behaviors and bodies, the range of its application will be expanded.
  We will continue our research while paying attention to the future trend of large-scale vision-language models.
}%
{%
  実験から得られた結果について考察する.
  全体を通して高い精度の状態認識器を得ることができたが, そこにはいくつかの特性が見て取れた.
  まず, 最適化を行うOPTのケースはほとんどの場合, 全てのプロンプトを用いるALLや最も良いプロンプトを一つだけ選ぶONEのケースよりも認識性能が高く, プロンプトの重み付け最適化の重要性が分かった.
  OPTのケースでは同じプロンプトにより多様なドアの開閉状態を認識でき, 透明なドアや綺麗汚いといった質的な状態の認識も可能である.
  対して, ALLとONEの性能差はタスクごとに異なり, 一概に判断することはできない.
  一方, ALLやONEの性能も簡単なタスクにおいては高く, 特にALLは重み付けの決定も必要ないため, タスクによってはALLを用いるのが最も容易である.

  次に, $R_{opt}$について, 最適化を行うOPTのケースではほとんどのタスクで100\%の認識率であった.
  一方, $R_{eval}$についてはその性能に差があり, (1)--(3)の3つの事が分かった.
  (1) OPT-1が単独でOPT-2やOPT-3よりも性能が高いことは無かった.
  (2) water experimentのケースを除き, 認識率が高いタスクの場合(例えばrefrigerator, various doors, light, display experiment)は, 大きな差ではないがOPT-3の方がOPT-2よりも認識率が高い.
  (3) 認識率が低いタスクの場合は(例えばcabinet, microwave, transparent door, handbag, kitchen experiment), OPT-2の方がOPT-3よりも認識率が高い.
  特に(2)(3)の事実について\figref{figure:discussion}を用いて考察する.
  これはそれぞれ$D_{opt}$の20枚の画像に対するopenまたはclosedの正解状態と, $e^{t}-C^{thre}$を, $e^{t}-C^{thre}$でソートした上で示している.
  $e^{t}-C^{thre}=0$の線から上をopenとして認識, 下をclosedとして認識している.
  ここで, \figref{figure:discussion}の上2つの図は(2)の事象, 下2つの図は(3)の事象に対応する.
  ElevatorとRefrigeratorの実験においては, 分散の最小化を考慮することでOPT-3の方がOPT-2に比べて綺麗に境界が分かれていることが分かる.
  これにより, 認識器がロバストになり, $R_{eval}$においてOPT-3がOPT-2の性能を上回っている.
  一方で, MicrowaveとTransparentの実験においては, 分散の最小化が上手く行かずに境界が作れていない, または正確に分類できない値が入ることで分散最小化がうまく働いていない.
  この場合, より平均値の差を最大化するOPT-2の方がOPT-3よりも高い性能を出している.
  なお, (2)における差は(3)ほど大きくないため, OPT-2を用いた場合が最も性能が安定していると考えられる.

  本研究の限界について述べる.
  本研究はほとんどのタスクについてOPT-2またはOPT-3により高い性能で状態認識が可能になることが分かった.
  その一方で, 特に水の認識の性能はまだ改善の余地があるし, その他認識についても$R_{eval}$の改善は今後行っていく必要がある.
  これらについてはVQAのようなタスクを使ったほうがより良い性能が得られる可能性があるが, 推論に時間を要するという問題もある\cite{kawaharazuka2023ofaga}.
  今後の大規模視覚-言語モデルの発展次第でもその性能は大きく向上すると考えられる.
  また, 今後本手法を用いたより実用的なシステム作りが必要である.
  環境の状態だけでなく, 食材の状態や身体の状態まで扱うことができれば, より適用範囲が広がるであろう.
  今後の大規模視覚-言語モデルの動向に注意しながら研究を進めたい.
}%

\section{CONCLUSION} \label{sec:conclusion}
\switchlanguage%
{%
  In this study, we described a binary state recognition method for robots through the spoken language using CLIP, a pre-trained large-scale vision-language model.
  By calculating the similarity between multiple prompts and the current image and combining them, the robot can recognize the state of objects and environments such as the open/closed state of a door or the on/off state of a light.
  By preparing a small number of datasets and optimizing the weighting of the prepared prompts using black-box optimization, a more accurate state recognizer can be constructed.
  We also discussed the change in recognition accuracy depending on the difference of evaluation functions in the optimization.
  Higher recognition accuracy can be obtained by maximizing the difference of the means of the evaluation values in two states or minimizing the variance of the evaluation value within each state.
  We have shown that the robot can easily recognize the open/closed state of a transparent door, whether water is running or not from a faucet, and even the qualitative state of whether a kitchen is clean or not, without manual programming, retraining of neural networks, or the use of dedicated sensors.
  Since only prompts and their weights need to be prepared for each recognizer, there is no need to prepare multiple models, which facilitates resource management.
  We believe that this study will revolutionize the recognition behavior of robots, and we intend to work on more advanced state recognition and related action generation in the future.
}%
{%
  本研究では事前学習済みの大規模視覚-言語モデルであるCLIPを用いたロボットのための二値状態認識手法について述べた.
  用意した複数のプロンプトと現在画像の類似度を計算し, これを統合することでドアの開閉や電気のオンオフなどの物体・環境の状態認識が可能となる.
  また, 少数のデータセットを用意し, これら類似度の重み付けを遺伝的アルゴリズムにより最適化することで, より高精度な状態認識器を構築することができる.
  加えて, 最適化における評価関数の違いに応じた認識精度の変化についても考察した.
  2つの状態間の評価値平均の差の最大化と, それぞれの状態内の評価値分散の最小化を行うことで, より高い認識精度が得られる.
  これらにより, 手動のプログラミングやニューラルネットワークの再学習, 専用のセンサを用いずとも, 蛇口からの出水状態や透明なドアの開閉状態, キッチンの綺麗さといった質的状態までもが容易に可能となることを示した.
  本研究はロボットの認識行動を大きく革新すると考えており, 今後はより高度な状態認識とそれに紐づく行動生成に取り組んでいきたい.
}%

{
  \bibliographystyle{IEEEtran}
  \bibliography{main}
}

\end{document}